\documentclass[letterpaper]{article} 
\usepackage{aaai2026}  
\nocopyright           
\usepackage{times}  
\usepackage{helvet}  
\usepackage{courier}  
\usepackage[hyphens]{url}  
\usepackage{graphicx} 
\usepackage{natbib}  
\usepackage{amsmath}
\usepackage{times}
\usepackage{latexsym}
\usepackage[T1]{fontenc}
\usepackage[utf8]{inputenc}
\usepackage{microtype}
\usepackage{inconsolata}

\usepackage{tikz}
\usepackage{algpseudocode}
\algrenewcommand\textproc{\text}
\usepackage{makecell}
\usepackage{booktabs}
\usepackage{color}
\usepackage{colortbl}
\usepackage{multirow}
\usepackage{enumitem}
\usepackage{makecell}
\usepackage{threeparttable}
\usepackage{amsmath,amsfonts,mathtools}
\usepackage{pifont}

\usepackage{mathrsfs}
\usepackage{graphicx}
\usepackage{xcolor}
\usepackage{enumitem}
\usepackage{tabularx, booktabs}

\usepackage{tcolorbox}

\definecolor{mygreen}{HTML}{00B050}

\definecolor{myorange}{HTML}{ED7D31}
\definecolor{lightgray}{gray}{0.95}
\definecolor{rowgray}{gray}{0.97}
\definecolor{avgblue}{RGB}{210,230,250}  
\definecolor{headergray}{RGB}{160,160,160} 

\usepackage{array}
\usepackage{mathtools}
\usepackage{multirow}
\usepackage{subcaption}
\usepackage{tikz}
\renewcommand{\arraystretch}{1.1}
\definecolor{color1}{cmyk}{0.216,0.176,0,0}
\definecolor{color2}{cmyk}{0.059,0.235,0.392,0}

\usepackage{graphicx}
\usepackage{tikz}
\usepackage{algorithm}
\usepackage{algpseudocode}
\algrenewcommand\textproc{\text}
\usepackage{makecell}
\usepackage{booktabs}
\usepackage{pifont}
\usepackage{multirow}
\usepackage{enumitem}
\usepackage{threeparttable}
\usepackage{amsmath,amsfonts,mathtools} 
\usepackage{longtable}
\usepackage{amsthm}

\usepackage{colortbl}
\usepackage{mathrsfs}
\usepackage{graphicx}
\usepackage{tabularx, booktabs}
\usepackage{tikz}

\definecolor{uc_color}{rgb}{0.99,0.24,0.63}
\definecolor{hc_color}{rgb}{0.02,0.51,0.51}
\definecolor{tc_color}{rgb}{0.99,0.55,0.09}

\tikzstyle{mybox} = [draw=black, very thick,
    rectangle, rounded corners, inner sep=10pt, inner ysep=13pt]
\tikzstyle{fancytitle} =[fill=black, text=white]

\usepackage{caption} 
\usepackage{hyperref}
\hypersetup{colorlinks=true, linkcolor=[rgb]{0.0,0.0,0.55}, citecolor=[rgb]{0.0,0.0,0.55}, urlcolor=[rgb]{0.0,0.0,0.55}}
\usepackage{newfloat}
\usepackage{listings}
\DeclareCaptionStyle{ruled}{labelfont=normalfont,labelsep=colon,strut=off} 
\floatstyle{ruled}
\newfloat{listing}{tb}{lst}{}
\floatname{listing}{Listing}
\title{\textit{KbSD}: Knowledge Boundary aware Self-Distillation for \\ Behavioral Calibration in Agentic Search}
\author{
    Tao Feng\textsuperscript{\rm 1}\equalcontrib,
    Xinke Jiang\textsuperscript{\rm 2}\equalcontrib,
    Chao Wu\textsuperscript{\rm 1}\thanks{Corresponding author.}
}
\affiliations{
    \textsuperscript{\rm 1}Zhejiang University, Hangzhou, China\\
    \textsuperscript{\rm 2}Peking University, Beijing, China
}

\begin{document}

\maketitle

\begin{abstract}
Agentic search equips large language models with dynamic retrieval abilities, but existing reinforcement learning methods remain limited by \textit{reward sparsity} in knowledge boundary calibration---deciding when to trust parametric memory, when to rely on retrieved evidence, and when to abstain.
Binary rewards can penalize undesirable outcomes, but provide little guidance on the reasoning process required to make calibrated decisions across different knowledge states.
To address, we propose \textbf{KbSD} (\textbf{K}nowledge \textbf{b}oundary \textbf{S}elf-\textbf{D}istillation), a framework that addresses this limitation through dense token-level supervision, outcome-level sparse rewards and quadrant-adaptive optimization.
KbSD constructs a hint-augmented \textit{teacher}, architecturally identical to the \textit{student}, that receives explicit knowledge boundary signals---including parametric certainty, retrieval quality, and ground-truth answers---to generate calibrated reasoning demonstrations.
This information-asymmetric self-distillation enables dense supervision without requiring a larger external model.
To further account for the heterogeneous reasoning distributions across knowledge states, we introduce a \textit{quadrant-adaptive} distillation objective: reverse KL for concentrated integration, forward KL for diverse refusal, and Pareto-optimal bidirectional KL for asymmetric quadrants requiring both precision and coverage.
Experiments on multiple benchmarks show that KbSD consistently improves both task accuracy and hallucination mitigation over strong baselines, with the largest gains appearing in the challenging quadrants where sparse rewards are least informative.
Our code is available at \href{https://github.com/jiangxinke/Agentic-RAG-R1/tree/kbsd}{this repository}.
\end{abstract}


\section{Introduction}
\textbf{L}arge \textbf{L}anguage \textbf{M}odels (\textbf{LLMs}) augmented with agentic search capabilities have achieved remarkable progress on knowledge-intensive reasoning tasks~\citep{yao2022react, jin2025search, nakano2021webgpt, singh2025agentic}. 
In this paradigm, models autonomously orchestrate queries, evaluate retrieved evidence, and iteratively refine their reasoning strategies through multi-turn interactions with external knowledge sources~\citep{dong2025arpo, dong2025aepo, li2025search}. 
A fundamental question in such agentic search systems is \textit{\textbf{knowledge boundary calibration}}: \textbf{determining when parametric memory is reliable, when external retrieval provides trustworthy evidence, and what strategy to adopt when these signals conflict or are both absent}.
\begin{figure}[t]
\centering
\includegraphics[width=\columnwidth]{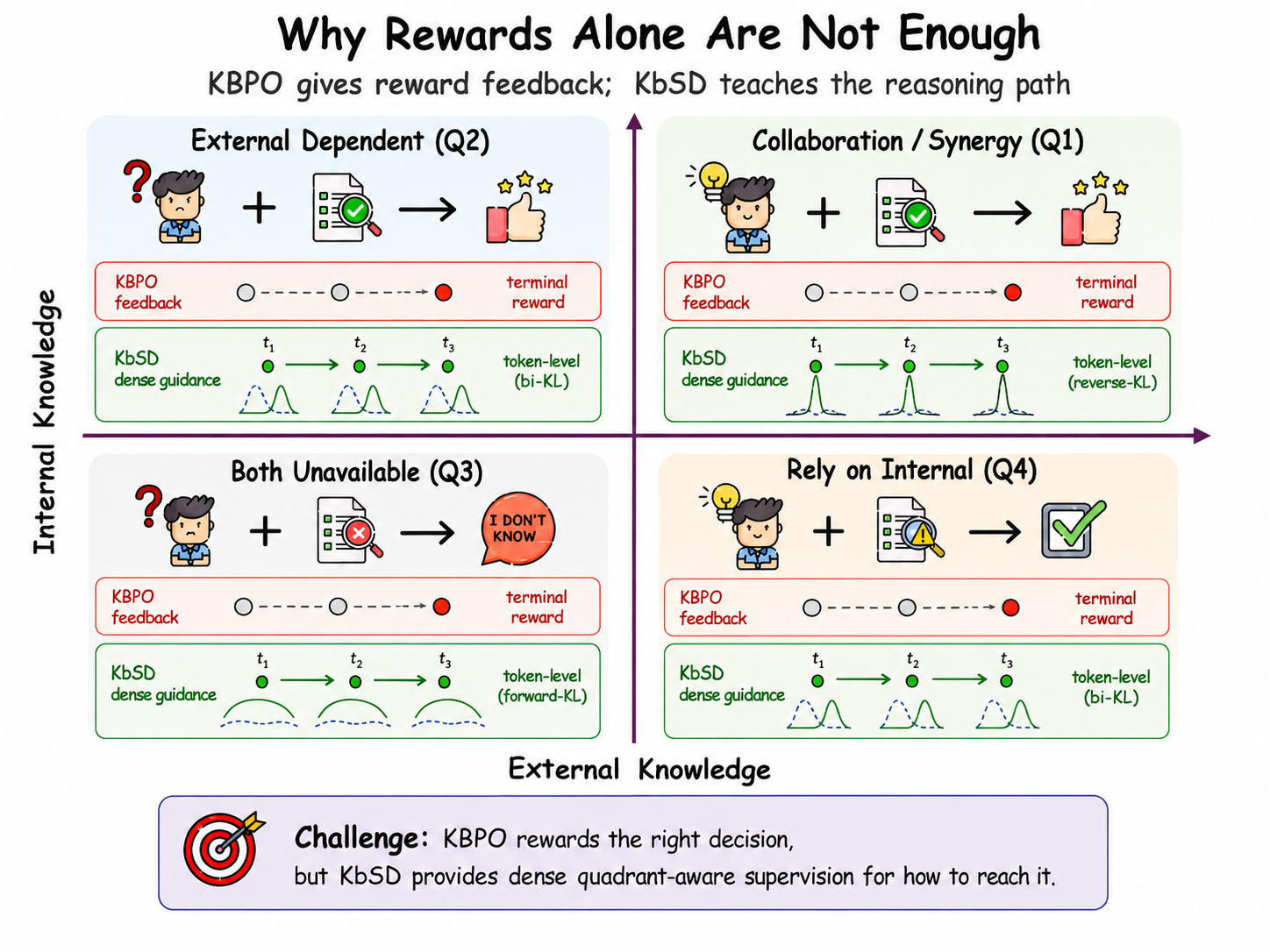}
\caption{KbPO vs.\ KbSD across the four knowledge-boundary quadrants. KbPO supplies only a terminal reward, whereas KbSD adds dense token-level supervision at each reasoning step with quadrant-adaptive KL objectives: reverse KL for Q1, bidirectional KL for Q2/Q4, and forward KL for Q3.}
\label{fig:intro}
\end{figure}
Traditional agentic search methods, such as Search-R1~\citep{jin2025search} and TC-RAG~\citep{jiang2024tcrag}, primarily focus on \textit{when} to \textbf{trigger retrieval}---that is, proactively searching when the model recognizes gaps in its own knowledge. 
However, the more critical question arises \textit{after} retrieval: does the retrieved evidence align with or contradict the model's parametric memory, and how should the agent reconcile these potentially conflicting knowledge sources? 
Recent work has begun to formalize this issue through cognitive taxonomies that characterize the interplay between internal certainty and retrieval quality. 
IKEA~\citep{huang2025reinforced}, CRaFT~\citep{zhu2025utilize}, and GRAIT~\citep{zhu2025grait} introduce a two-quadrant framework that distinguishes between known and unknown queries, guiding the model to search only when parametric knowledge is insufficient. 
KbPO~\citep{feng2025kbpo} further refines this into a four-quadrant taxonomy by jointly considering internal knowledge reliability and retrieval quality: the agent should integrate, rely on external knowledge, rely on internal knowledge, or refuse to answer under the four corresponding situations (Figure~1). 
These frameworks operationalize such taxonomies through quadrant-aware loss or reward during post-training, enabling more targeted calibration of retrieval-augmented behaviors.

Despite these advances, existing approaches share a fundamental bottleneck: \textit{reward sparsity at the process level}. 
In knowledge-boundary-aware frameworks, the central objective is to guide the agent toward selecting the most appropriate behavior during search. 
Yet the combination of sparse reward signals and the large search space induced by four distinct behavioral modes means that the agent mostly receives negative feedback. 
A binary reward merely indicates \textit{what not to do}---penalizing a hallucination or a redundant search---without revealing \textit{which quadrant strategy should have been adopted} or \textit{how the agent should have reasoned toward that decision}. 
Consider a Quadrant~3 scenario: the agent must refuse to answer a question about an obscure entity when retrieval returns irrelevant results. 
A sparse reward can penalize hallucination \textit{post hoc}, but it cannot decompose the cognitive steps that \textbf{distinguish principled abstention from uninformed guessing}, such as assessing retrieval relevance, gauging internal uncertainty, and reasoning toward refusal. 
Addressing this reward sparsity problem raises two fundamental challenges:

\textbf{\textit{Challenge~1: From Sparse Rewards to Dense Supervision.}} 
Sparse rewards operate only at action boundaries, providing a scalar judgment on the final outcome while leaving the intermediate reasoning trajectory entirely unsupervised. 
To bridge this gap, the agent requires dense, token-level guidance that demonstrates not only \textit{which} quadrant strategy to adopt, but also the full cognitive process leading to that decision: how to assess internal certainty, evaluate retrieved evidence, and synthesize these signals into a calibrated conclusion.

\textbf{\textit{Challenge~2: Quadrant-Heterogeneous Optimization Geometry.}} 
Even with dense supervision, a uniform training objective is ill-suited to the four knowledge states, which exhibit fundamentally different distributional characteristics in their reasoning patterns. 
In Quadrant~1, where both sources are reliable and mutually consistent, the correct reasoning process is concentrated and relatively unimodal---there is typically one clear way to integrate concordant evidence. 
In Quadrant~3, by contrast, refusal behavior is inherently diverse: the agent should express uncertainty through varied reasoning paths rather than collapse to a single template. 
Quadrants~2 and~4 require both precision in source selection and robustness to diverse forms of retrieval noise. 
One-size-fits-all divergence measure therefore risks over-covering concentrated distributions or under-covering diverse ones, leading to suboptimal calibration.

To address, we propose \textbf{K}nowledge \textbf{B}oundary aware \textbf{S}elf-\textbf{D}istillation \textbf{(KbSD)}, a quadrant-adaptive self-distillation framework for knowledge boundary calibration in agentic search. 
Our approach first quantifies the model's parametric knowledge boundary by probing data certainty and semantic stability, partitioning training queries into distinct knowledge states that define the behavioral norm required for each query. 
These boundary signals, together with retrieval quality assessments and ground-truth answers, are injected as \textit{knowledge boundary hints} into a copy model, forming a hint-augmented \textit{teacher} that shares identical parameters with the hint-free \textit{student}. 
This information-asymmetric design provides dense process-level demonstrations: the teacher reasons with full boundary awareness, while the student learns to internalize these calibrated reasoning patterns through token-level supervision (Challenge~1). 
To further address the distributional heterogeneity across quadrants (Challenge~2), we design a quadrant-adaptive KL objective: reverse KL (mode-seeking) for Quadrant~1's concentrated integration, forward KL (mass-covering) for Quadrant~3's diverse refusal patterns, and Pareto-optimal bidirectional KL for the competing precision--coverage trade-off in Quadrants~2 and~4. 
This design is integrated with GRPO-based policy optimization, yielding complementary dense supervision and sparse reward signals.

In summary, our contributions are as follows:
\begin{itemize}
\item We identify reward sparsity as a fundamental bottleneck in knowledge boundary calibration and propose a self-distillation framework that leverages \textit{information asymmetry}---knowledge boundary hints accessible only to the teacher---to provide dense process-level supervision without requiring a larger or separately trained model.
\item We design a quadrant-adaptive distillation strategy that matches optimization geometry to each knowledge state: reverse KL for Q1, forward KL for Q3, and Pareto-optimal multi-objective optimization over both divergence directions for Q2 and Q4.
\item Extensive experiments across multiple benchmarks demonstrate that our approach significantly reduces hallucination rates and improves calibration quality over strong baselines, with particularly pronounced gains on the challenging quadrants where sparse rewards are least effective.
\end{itemize}

\section{Related Work for Agentic Search}
\label{sec:related_work}

\begin{figure*}[t]
\centering
\includegraphics[width=\textwidth]{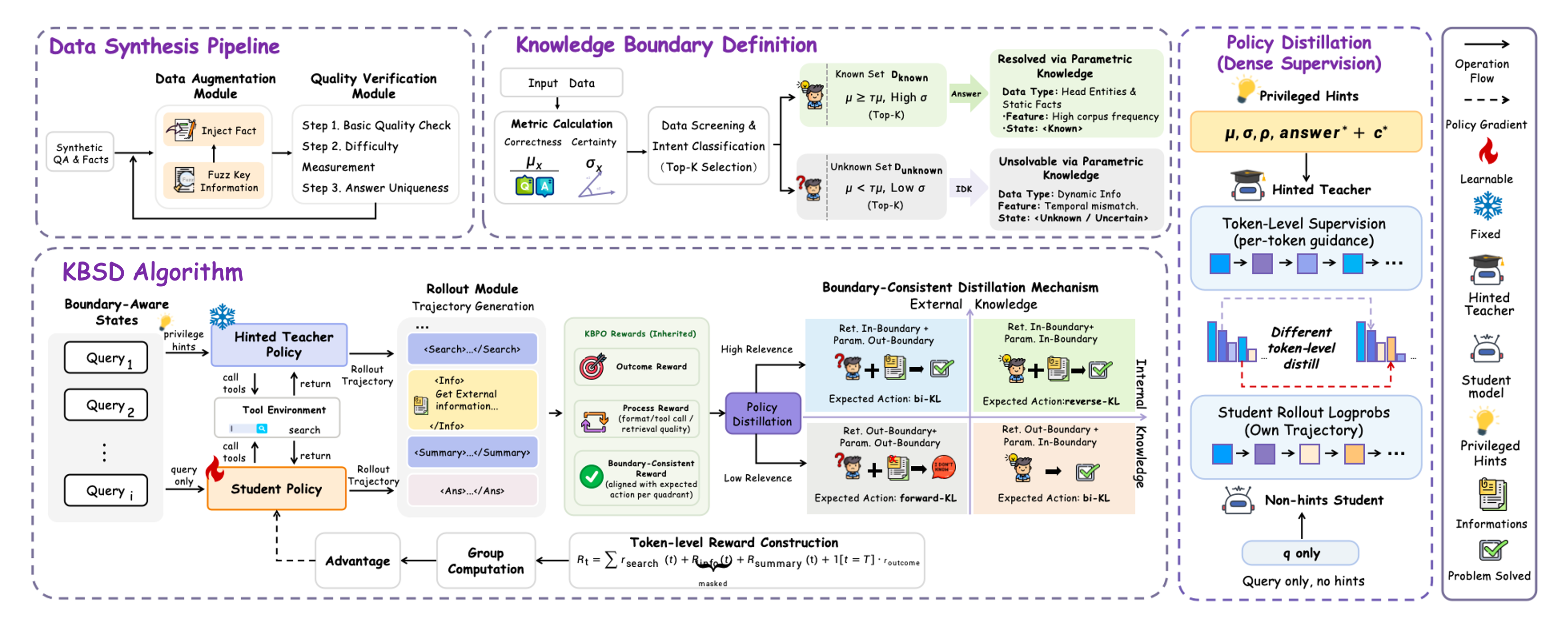}
\caption{
Overview of \textsc{KbSD}. We first synthesize boundary-challenging queries, then estimate each query's knowledge state using parametric certainty, semantic stability, and retrieval quality, and assign it to one of four behavioral quadrants. Based on these signals, a privileged hinted teacher generates calibrated reasoning demonstrations, which are distilled into a hint-free student with quadrant-adaptive objectives jointly optimized with GRPO. In this way, \textsc{KbSD} combines dense process-level supervision with sparse reward optimization.
}
\label{fig:workflow}
\end{figure*}

Retrieval-Augmented Generation (RAG) grounds LLM outputs in external knowledge, and later extensions improve it with more adaptive retrieval and reasoning strategies~\citep{lewis2020retrieval, guu2020retrieval, jiang2023active, asai2024self, yan2024corrective, trivedi2023interleaving}.
Agentic search further treats retrieval as a dynamic decision-making process: ReAct~\citep{yao2023react} interleaves reasoning traces with tool-use actions to enable on-the-fly query formulation, WebGPT~\citep{nakano2021webgpt} trains agents to navigate web search through human feedback, and TC-RAG~\citep{jiang2024tcrag} equips agents with stack memory to maintain coherent reasoning across multi-turn retrieval trajectories.
Recent work applies reinforcement learning to optimize such search agents---Search-R1~\citep{jin2025search} leverage GRPO~\citep{shao2024deepseekmath} to train retrieval strategies via trajectory-level outcome rewards, ARPO~\citep{dong2025arpo} and AEPO~\citep{dong2025aepo} introduce entropy-based adaptive rollouts to improve exploration efficiency, and MEM1~\citep{zhou2025mem1} augments agents with explicit memory modules for retaining intermediate findings.

\subsection{Knowledge Boundary Calibration}
These agentic methods optimize for final-answer correctness without explicitly modeling the reliability of the model's internal knowledge or the quality of retrieved evidence.
Recent studies therefore begin to explore \textit{knowledge boundary calibration} for agentic search.
IKEA~\citep{huang2025reinforced} models a known-unknown distinction to penalize redundant retrieval for queries the model can already answer, while KbPO~\citep{feng2025kbpo} extends this to a four-quadrant formulation that jointly considers internal certainty and retrieval quality, designing quadrant-specific alignment rewards within the GRPO framework to incentivize integration, reliance, refusal, or parametric trust as appropriate.
\textbf{However, existing methods still rely on sparse reward or loss signals that specify the desired behavioral mode without supervising the intermediate reasoning process that leads to it---a gap our self-distillation framework directly addresses.}

\subsection{Knowledge-Distillation}
Knowledge distillation (KD)~\citep{hinton2015distilling} transfers capabilities from a large teacher model into a compact student.
Classical off-policy KD trains students on teacher-generated sequences, which introduces \textit{exposure bias}: at inference the student must condition on its own outputs rather than the teacher's ideal prefixes, causing distributional mismatch.
On-Policy Distillation (OPD) eliminates this mismatch by supervising students on their own rollouts.
MiniLLM~\citep{gu2024minillm} replaces forward KL with reverse KL to achieve mode-seeking behavior on student-sampled sequences; GKD~\citep{agarwal2024policy} provides a unified on-policy token-level matching framework for learning from self-generated mistakes; and DistiLLM-2~\citep{ko2025distillm} introduces a contrastive objective that treats teacher- and student-generated outputs asymmetrically.
Recent work removes the separate teacher entirely: Self-Distillation Zero~\citep{he2026self} converts sparse binary rewards into dense process-level supervision via self-revision, and Self-Distilled RLVR~\citep{yang2026self} integrates self-distillation with RL training.
\citet{yang2026learning} and \citet{li2026rethinking} further analyze OPD dynamics and propose reward-augmented extensions.
\textbf{However, existing OPD approaches either depend on a separately trained teacher of larger capacity, or apply a uniform divergence objective across all training instances---neither accounts for the heterogeneous reasoning distributions that arise from distinct knowledge boundary states.}

\section{Method}
\label{sec:method}

Figure~\ref{fig:workflow} illustrates the overall pipeline of \textsc{KbSD}, a knowledge-boundary-aware self-distillation framework for agentic retrieval.
Existing boundary-aware reinforcement learning methods can indicate \emph{which} behavior is desirable under different knowledge states---e.g., trust parametric memory, rely on retrieval, integrate both sources, or refuse---but they typically provide only sparse trajectory-level supervision.
\textsc{KbSD} addresses this limitation by converting coarse behavioral prescriptions into \textbf{dense process-level supervision}.
Our framework has three stages: \ding{182} knowledge-boundary-aware construction of challenging multi-hop queries; \ding{183} reliability and evidence sufficiency estimation with boundary-quadrant mapping; \ding{184} hint-guided teacher--student distillation with quadrant-adaptive reinforcement learning.

\subsection{Boundary-Oriented Data Synthesis}
\label{sec:data_evolution}

To train the model on queries near or beyond its parametric knowledge boundary, we synthesize challenging long-horizon questions from simple seed QA examples.
Following prior work~\citep{feng2025kbpo,gao2025beyond}, we begin with a seed corpus
$\mathcal{D}_0 = \{(q, a, \mathcal{F})\}$,
where $q$ denotes a question, $a$ is its answer, and $\mathcal{F}$ its supporting facts.
Each sample is iteratively transformed for $T$ rounds using two complementary operators named \textit{\textbf{Fact Injection}} and \textit{\textbf{Information Fuzzing}}:
\begin{itemize}
    \item \textbf{\textit{Fact Injection.}} The first operator, $\mathcal{O}_{\mathrm{inj}}$, increases compositional depth by introducing an additional fact hop. Given a question $q$ and a target entity $e \in q$, we retrieve a new fact $f_{\mathrm{new}}$ associated with $e$ and rewrite the question so that solving it requires combining the original evidence with the injected fact: \begin{equation}
(q', \mathcal{F}') = \mathcal{O}_{\mathrm{inj}}(q, \mathcal{F}, e),
\quad
\mathcal{F}' = \mathcal{F} \cup \{f_{\mathrm{new}}\}.
\end{equation}This operation makes the reasoning chain longer and pushes the query away from direct memorization.
    \item \textbf{\textit{Information Fuzzing.}}
The second operator, $\mathcal{O}_{\mathrm{fuzz}}$, weakens explicit constraints by replacing a precise condition $c$ with an underspecified or indirect description $\tilde{c}$:
\begin{equation}
q' = \mathcal{O}_{\mathrm{fuzz}}(q, c \rightarrow \tilde{c}).
\end{equation}
This transformation preserves answerability while increasing ambiguity and retrieval difficulty.
\end{itemize}
After synthesis, each sample is filtered through multi-stage validation to ensure semantic validity, sufficient difficulty, and answer uniqueness~\citep{gao2025beyond}.
The resulting dataset is more reliably expose failure modes in both internal knowledge and retrieval-based reasoning.

\subsection{Knowledge Boundary Quantification}
\label{sec:boundary_quantification}
To quantify the knowledge-boundary state of each query, we jointly model two complementary aspects: the reliability of the model's internal knowledge and the sufficiency of external retrieved evidence. 

\paragraph{Internal Knowledge Quantification.} For each query $q$, we assess internal knowledge reliability using two complementary  signals derived from retrieval-free responses:
\begin{itemize}
    \item {\textit{Parametric Certainty.}}
We first measure whether the model can answer $q$ from internal knowledge alone.
Using the frozen base model without retrieval, we sample $N$ independent responses $\{y_i\}_{i=1}^{N}$ and compute the fraction that matches the ground-truth answer via $\mathbf{I}(\cdot)$:
\begin{equation}
\mu(q) = \frac{1}{N} \sum_{i=1}^{N} \mathbf{I}\!\left[\operatorname{Match}(y_i, a)\right].
\end{equation}
A high value of $\mu(q)$ indicates that the answer is likely contained in the model's parametric memory.

\item {\textit{Semantic Stability.}}
Correctness alone does not reveal whether the model's internal knowledge is stable or brittle.
We therefore compute the average pairwise semantic similarity among sampled responses:
\begin{equation}
\sigma(q) =
\frac{2}{N(N-1)}
\sum_{1 \le i < j \le N}
\cos\!\left(\operatorname{Enc}(y_i), \operatorname{Enc}(y_j)\right),
\end{equation}
where high $\sigma(q)$ suggests that the model converges to a consistent latent belief and low $\sigma(q)$ indicates uncertainty or internal inconsistency.
\end{itemize}


\paragraph{External Knowledge Quantification.}
In addition to internal knowledge, we estimate whether retrieval provides usable support. Let $\hat{\rho}(q)$ denote a retrieval-quality score computed from the evidence returned for $q$. A higher $\hat{\rho}(q)$ indicates that the retrieved evidence is more likely to support answer generation.

\paragraph{Boundary-Quadrant Mapping.}
We next combine the internal and external signals to determine the knowledge-boundary state of each query. Specifically, we binarize internal knowledge reliability and retrieval sufficiency as
\begin{equation}
k(q) = \mathbf{I}\!\left[\mu(q) \ge \tau_{\mu}\right],
\quad
s(q) = \mathbf{I}\!\left[\hat{\rho}(q) \ge \tau_{\rho}\right].
\end{equation}
These two binary indicators determine the desirable source-allocation behavior for query $q$: the model should integrate internal knowledge with retrieved evidence when both are reliable, rely on retrieval when internal knowledge is insufficient, trust internal knowledge when retrieval is weak, and abstain when neither source is reliable. Formally, we map the pair $(s(q), k(q))$ to one of four behavioral quadrants:
\begin{equation}
c^{*}(q) =
\begin{cases}
\text{Integrated}, & s(q)=1,\; k(q)=1,\\
\text{External},   & s(q)=1,\; k(q)=0,\\
\text{Refusal},    & s(q)=0,\; k(q)=0,\\
\text{Internal},   & s(q)=0,\; k(q)=1.
\end{cases}
\end{equation}
We further use $\sigma(q)$ as a stability filter in training-set construction, rather than incorporating it directly into quadrant assignment. Concretely, we retain high-stability samples in the known region and low-stability samples in the unknown region, yielding a sharper boundary partition.
\subsection{Hint-Augmented Self-Distillation}
\label{sec:self_distillation}

The central idea of \textsc{KbSD} is to transform boundary annotations into dense reasoning supervision.
Rather than rewarding only the final behavioral mode, we construct a \emph{privileged teacher} that has access to boundary hints during training and use it to generate calibrated reasoning demonstrations.
A \emph{student} without such hints is then trained to reproduce these behaviors from the original input alone.

Specifically, for each query $q$, we build a structured hint:
\begin{equation}
h(q) = \big(\mu(q),\, \sigma(q),\, \hat{\rho}(q),\, c^{*}(q)\big),
\end{equation}
which summarizes the model's internal knowledge state, response stability, retrieval reliability, and the target behavioral regime.
We prepend this hint to the original query to form the teacher input $p^{T}(q) = [h(q);\, q].$
The teacher policy is defined as the same underlying model conditioned on the privileged prompt:
\begin{equation}
\pi^{T}(\cdot \mid q) =\operatorname{stopgrad}\!\big(\pi_{\theta}(\cdot \mid p^{T}(q))\big).
\end{equation}
Thus, the teacher and student share the same architecture and initialization, but only the teacher has access to boundary hints; gradients are not propagated through the teacher branch.
This information asymmetry allows the teacher to produce reasoning trajectories that are better calibrated to the query's knowledge state, while the student must infer the appropriate behavior without privileged signals at test time.

The hints are used only during training and are never provided at inference time.
Their purpose is not to replace answer supervision, but to make the latent behavioral target explicit.
For example, a query in the \textit{Refusal} quadrant should demonstrate evidence inspection and uncertainty-aware abstention, whereas a query in the \textit{Integrated} quadrant should demonstrate how internal knowledge and retrieved evidence are reconciled.
By conditioning on boundary hints, the teacher turns a sparse mode label into a complete reasoning trajectory that can supervise the student token by token.

\subsection{Quadrant-Adaptive Distillation}
\label{sec:adaptive_distillation}

Different knowledge-boundary states induce qualitatively different target distributions over valid reasoning trajectories.
We therefore avoid using a uniform distillation objective for all samples.
Instead, we select the distillation direction according to the structure of the target behavior in each quadrant.
Let $o^{T} \sim \pi^{T}(\cdot \mid q)$ denote a teacher-generated trajectory and $o \sim \pi_{\theta}(\cdot \mid q)$ a student rollout.
We consider two token-level distillation objectives.

\paragraph{Forward Distillation.}
The first objective imitates teacher trajectories by maximizing their likelihood under the student:
\begin{equation}
\mathcal{L}_{\mathrm{fwd}}(q)
=
-\mathbb{E}_{o^{T} \sim \pi^{T}}
\left[
\sum_{t=1}^{|o^{T}|}
\log \pi_{\theta}(o^{T}_{t} \mid o^{T}_{<t}, q)
\right].
\label{eq:fwd_kl_rewrite}
\end{equation}
This objective is mass-covering: it encourages the student to support the full set of teacher behaviors.

\paragraph{Reverse Distillation.}
The second objective matches the teacher on student-sampled trajectories:
\begin{equation}
\mathcal{L}_{\mathrm{rev}}(q)
=
\mathbb{E}_{o \sim \pi_{\theta}}
\left[
\sum_{t=1}^{|o|}
\log
\frac{\pi_{\theta}(o_t \mid o_{<t}, q)}
{\pi^{T}(o_t \mid o_{<t}, q)}
\right].
\label{eq:rev_kl_rewrite}
\end{equation}
This objective is mode-seeking: it suppresses student behaviors unsupported by the teacher and sharpens the policy around dominant teacher modes.

\paragraph{Pareto Bidirectional Distillation.}
For the \textit{External} and \textit{Internal} quadrants, we combine forward and reverse distillation through a Pareto-style adaptive weighting.
Let $
\mathbf{g}_{f} = \nabla_{\theta} \mathcal{L}_{\mathrm{fwd}}(q),
\mathbf{g}_{r} = \nabla_{\theta} \mathcal{L}_{\mathrm{rev}}(q).
$
We seek a common descent direction by solving
\begin{equation}
\alpha^{*}
=
\arg\min_{\alpha \in [0,1]}
\left\|
\alpha \mathbf{g}_{f} + (1-\alpha)\mathbf{g}_{r}
\right\|^{2},
\label{eq:mgda_rewrite}
\end{equation}
which has the closed-form solution
\begin{equation}
\alpha^{*}
=
\operatorname{clip}
\left(
\frac{(\mathbf{g}_{r}-\mathbf{g}_{f})^{\top}\mathbf{g}_{r}}
{\|\mathbf{g}_{f}-\mathbf{g}_{r}\|^{2}},
0,1
\right).
\end{equation}
The resulting objective is
\begin{equation}
\mathcal{L}_{\mathrm{pareto}}(q)
=
\alpha^{*}\mathcal{L}_{\mathrm{fwd}}(q)
+
(1-\alpha^{*})\mathcal{L}_{\mathrm{rev}}(q).
\label{eq:pareto_rewrite}
\end{equation}
This adaptive combination removes the need to hand-tune a global trade-off coefficient and automatically balances coverage and precision according to the local gradient geometry.

\paragraph{Quadrant-Specific Assignment.}
We assign these objectives according to the expected diversity of valid reasoning in each quadrant:
\begin{equation}
\mathcal{L}_{\mathrm{distill}}(q)=
\begin{cases}
\mathcal{L}_{\mathrm{rev}}(q),
& c^{*}(q)=\text{Integrated},\\[3pt]
\mathcal{L}_{\mathrm{pareto}}(q),
& c^{*}(q)\in\{\text{External}, \text{Internal}\},\\[3pt]
\mathcal{L}_{\mathrm{fwd}}(q),
& c^{*}(q)=\text{Refusal}. \notag
\end{cases}
\label{eq:quadrant_distill_rewrite}
\end{equation}
The intuition is as follows:
\begin{itemize}
    \item In the \textit{Integrated} quadrant, both internal knowledge and retrieved evidence are reliable, so correct reasoning is typically concentrated around a relatively consistent integration pattern; a mode-seeking objective is therefore preferable.
    \item In the \textit{Refusal} quadrant, by contrast, there may be multiple acceptable ways to inspect evidence, recognize insufficiency, and articulate abstention; a mass-covering objective better preserves this diversity.
    \item The \textit{External} and \textit{Internal} quadrants require both precise source selection and robustness to varied noise patterns, making neither pure objective sufficient on its own.
\end{itemize}

\subsection{Joint RL and Distillation Training}
\label{sec:joint_training}

The distillation objective provides dense process-level supervision, but by itself it does not optimize final task performance.
We therefore combine it with GRPO-based reinforcement learning so that the student is trained both to \emph{behave appropriately} and to \emph{solve the task correctly}.

\paragraph{Reward Design.}
Following KbPO, for each rollout $o_i$ of length $T_i$, we define a composite reward at step $t$ as
\begin{equation}
r_{i,t}
=
\mathbf{I}[t=T_i] \cdot r_{i}^{\mathrm{out}}
+
\lambda r_{i,t}^{\mathrm{proc}}
+
r_{i,t}^{\mathrm{align}}
+
r_{i}^{\mathrm{pen}},
\end{equation}
where $r_{i}^{\mathrm{out}}$ is the final-answer reward (e.g., F1), $r_{i,t}^{\mathrm{proc}}$ rewards valid intermediate tool use, and $r_{i,t}^{\mathrm{align}}$ encourages behavior consistent with the target quadrant $c^{*}(q)$.
To mitigate over-refusal, we additionally impose a refusal penalty on answerable queries:
\begin{equation}
r_{i}^{\mathrm{pen}}=
\begin{cases}
-\phi,
& \text{if } c^{*}(q_i)\in\{\textit{Integrated}, \textit{External}, \textit{Internal}\} \\
& \quad \text{and the model refuses},\\
0, & \text{otherwise}. \notag
\end{cases}
\end{equation}

\paragraph{Training Objective.}
For each query $q$, the student policy generates a group of $G$ rollouts $\{o_i\}_{i=1}^{G}$.
We optimize the following joint objective:
\begin{equation}
\small
\begin{aligned}
\mathcal{L}(\theta)
=
-\mathbb{E}_{q \sim \mathcal{D}}
\Bigg[
&\frac{1}{G}\sum_{i=1}^{G}
\min\!\Big(
\rho_i \hat{A}_i,\,
\operatorname{clip}(\rho_i, 1-\epsilon, 1+\epsilon)\hat{A}_i
\Big) \\
&\quad - \beta\, \mathbb{D}_{\mathrm{KL}}(\pi_{\theta}\|\pi_{\mathrm{ref}})
- \gamma\, \mathcal{L}_{\mathrm{distill}}(q)
\Bigg], \notag
\end{aligned}
\label{eq:joint_objective_rewrite}
\end{equation}
where $
\rho_i = \frac{\pi_{\theta}(o_i \mid q)}{\pi_{\theta_{\mathrm{old}}}(o_i \mid q)},
\hat{A}_i = \frac{R_i - \operatorname{mean}(\mathbf{R})}{\operatorname{std}(\mathbf{R})}.
$
Here, $\pi_{\mathrm{ref}}$ is the reference policy used for KL regularization, and $\gamma$ controls the strength of distillation.
The two components play complementary roles.
The RL term supplies task-level optimization from \textbf{sparse rewards}, while the distillation term provides \textbf{dense, boundary-aware process supervision} derived from the hinted teacher.
Together, they train a retrieval agent that not only improves final-answer accuracy, but also learns \emph{how} to calibrate trust between internal knowledge and external evidence under different knowledge-boundary states.

\section{Experiment}

\subsection{Experimental Setup}

\paragraph{Datasets.}
We construct our training set using the data synthesis pipeline (Section~\ref{sec:data_evolution}) based on \textbf{HotpotQA}~\citep{yang2018hotpotqa} and \textbf{2WikiMultiHopQA}~\citep{ho2020constructing}. For evaluation, we use 9 benchmarks spanning three categories:
(1) \textit{Multi-hop QA}: HotpotQA, 2WikiMultiHopQA, MuSiQue~\citep{trivedi2022musique}, and Bamboogle~\citep{press2023measuring};
(2) \textit{Open-domain QA}: NQ~\citep{kwiatkowski2019natural}, TriviaQA~\citep{joshi2017triviaqa}, and PopQA~\citep{mallen2023popqa};
(3) \textit{Agentic reasoning}: FRAMES~\citep{krishna2025fact}, GAIA~\citep{mialon2023gaia}.

\paragraph{Models and Retrieval.}
We use \textbf{Qwen2.5}~\citep{Yang2024Qwen25TR} at two scales (3B, 7B) as backbone models. For retrieval, we use E5~\citep{wang2022text} as the dense retriever over the 2018 Wikipedia dump~\citep{karpukhin2020dense}, retrieving top-3 passages per query.

\paragraph{Training.}
We implement KbSD based on the VERL framework with GRPO as the optimizer. For each query, the policy generates $G{=}5$ rollouts with batch size 128 and 8 rollouts per batch. We train for 300 steps with learning rate $1{\times}10^{-6}$, KL coefficient $\beta{=}0.001$, process reward weight $\lambda{=}0.2$, and distillation weight $\gamma{=}0.1$. The knowledge boundary thresholds are set to $\tau_\rho{=}0.5$ and $\tau_\mu{=}0.6$. The agent operates in multi-turn mode with a maximum of 2 search iterations per query.

\paragraph{Baselines.}
We compare against three categories of methods:
(1) \textit{Standard RAG}: Naive RAG~\citep{lewis2020retrieval}, FS-RAG~\citep{trivedi2023interleaving}, and FLARE~\citep{jiang2023active};
(2) \textit{Agentic RAG}: ReAct~\citep{yao2023react}, IRCoT~\citep{trivedi2023interleaving}, Self-RAG~\citep{asai2024self}, and CRAG~\citep{yan2024corrective};
(3) \textit{RL-based Agents}: Search-R1~\citep{jin2025search}, ARPO~\citep{dong2025arpo}, AEPO~\citep{dong2025aepo}, MEM1~\citep{zhou2025mem1}, and KbPO~\citep{feng2025kbpo}.

\paragraph{Evaluation Metrics.}
We assess performance on two dimensions:
(1) \textbf{Task Utility}: \textbf{F1} scores measuring answer correctness;
(2) \textbf{Unreliability Rate}: defined as $1 - (\text{F1} + \text{Refusal Rate})$, quantifying ``unsafe'' behaviors where the agent neither answers correctly nor refuses faithfully. Lower is better.

\subsection{Main Results}

\begin{table*}[!ht]
\centering
\fontsize{7pt}{8pt}\selectfont
\setlength{\tabcolsep}{4.6pt}
\renewcommand{\arraystretch}{0.85}
\resizebox{\textwidth}{!}{
\begin{tabular}{l | l | c c | c c c c c c c | c}
\toprule
\rowcolor{gray!30}
\multicolumn{2}{c|}{\textbf{Method}} &
\multicolumn{2}{c|}{\textbf{In-Domain}} &
\multicolumn{7}{c|}{\textbf{Out-of-Domain}} &
\textbf{Avg} \\

\rowcolor{gray!30}
\textbf{Paradigm} & \textbf{Approach} &
\textbf{2Wiki} & \textbf{HotpotQA} &
\textbf{Bamboogle} & \textbf{FRAMES} & \textbf{GAIA} &
\textbf{MusiQue} & \textbf{NQ} & \textbf{PopQA} & \textbf{TriviaQA} &
\textbf{Avg} \\

\midrule

\multicolumn{12}{c}{\textbf{\textit{Qwen2.5-3B}}} \\
\midrule

\rowcolor{rowgray}\multirow{2}{*}{No RAG}
& Base & 76.02 & 75.92 & 90.55 & 91.99 & 94.30 & 90.30 & 85.73 & 87.89 & 61.16 & 83.76 \\
& COT & 81.10 & 76.18 & 79.20 & 92.84 & 92.98 & 89.53 & 83.47 & 88.63 & 60.38 & 82.70 \\
\midrule

\rowcolor{rowgray}\multirow{2}{*}{Naive RAG}
& FS-RAG & 84.53 & 74.15 & 89.52 & 89.58 & 95.70 & 92.36 & 80.16 & 69.59 & 54.62 & 81.13 \\
& FL-RAG & 83.20 & 73.22 & 88.95 & 90.81 & 98.42 & 92.71 & 78.07 & 66.53 & 51.50 & 80.38 \\
\midrule

\rowcolor{rowgray}\multirow{3}{*}{Agentic RAG}
& ReACT & 74.91 & 65.63 & 75.14 & 89.47 & 95.22 & 86.08 & 72.81 & 63.53 & 53.96 & 75.19 \\
& IRCOT & 84.11 & 75.50 & 74.73 & 93.21 & 97.23 & 87.57 & 72.14 & 64.98 & 50.81 & 77.81 \\
\rowcolor{rowgray}& TCRAG & 71.53 & 78.06 & 82.41 & 92.31 & 94.47 & 91.01 & 79.26 & 82.31 & 49.54 & 80.10 \\
\midrule

& ReSearch & 72.77 & 66.04 & 84.91 & 90.00 & 95.52 & 90.53 & 65.39 & 64.03 & 46.07 & 75.03 \\
\rowcolor{rowgray}& Search-R1 & 70.10 & 62.76 & 70.10 & 89.24 & 95.32 & 86.47 & 65.27 & 65.76 & 44.92 & 72.22 \\
& AEPO & 76.99 & 71.29 & 77.91 & 87.74 & 94.99 & 88.30 & 73.24 & 70.52 & 52.22 & 77.02 \\
\rowcolor{rowgray}& ARPO & 70.45 & 63.52 & 72.68 & 86.51 & 92.29 & 86.62 & 66.71 & 66.56 & 46.34 & 72.41 \\
& Mem1 & 81.94 & 79.85 & 94.81 & 95.01 & 97.53 & 95.53 & 80.82 & 83.29 & 66.91 & 86.19 \\
\rowcolor{rowgray}& KbPO & 70.50 & 57.60 & 68.10 & 62.60 & 77.20 & 61.40 & 68.50 & 73.10 & 39.90 & 64.32 \\
\rowcolor{avgblue}\multirow{-7}{*}{\makecell{RL-based\\Agentic RAG}}& \textbf{KbSD (Ours)} & \textbf{45.90} & \textbf{41.60} & \textbf{52.20} & \textbf{44.00} & \textbf{58.10} & \textbf{42.10} & \textbf{64.40} & \textbf{45.60} & \textbf{34.80} & \textbf{47.60} \\
\midrule

\multicolumn{12}{c}{\textbf{\textit{Qwen2.5-7B}}} \\
\midrule

\rowcolor{rowgray}\multirow{2}{*}{No RAG}
& Base & 74.59 & 73.37 & 82.14 & 87.48 & 95.71 & 87.85 & 80.28 & 85.21 & 50.92 & 79.73 \\
& COT & 76.45 & 70.90 & 62.44 & 82.40 & 93.83 & 85.65 & 77.53 & 84.93 & 50.67 & 76.09 \\
\midrule

\rowcolor{rowgray}\multirow{2}{*}{Naive RAG}
& FS-RAG & 82.29 & 70.79 & 83.14 & 87.48 & 94.98 & 89.26 & 83.18 & 81.06 & 64.98 & 81.91 \\
& FL-RAG & 80.22 & 65.58 & 75.90 & 87.90 & 95.40 & 87.54 & 80.28 & 74.72 & 57.34 & 78.32 \\
\midrule

\rowcolor{rowgray}\multirow{3}{*}{Agentic RAG}
& ReACT & 72.49 & 57.19 & 72.37 & 84.71 & 93.63 & 80.66 & 69.99 & 63.71 & 45.45 & 71.13 \\
& IRCOT & 63.55 & 73.71 & 78.10 & 93.22 & 94.50 & 91.61 & 80.37 & 65.86 & 50.57 & 76.83 \\
\rowcolor{rowgray}& TCRAG & 70.30 & 59.17 & 74.87 & 83.54 & 91.79 & 82.44 & 70.99 & 62.83 & 45.22 & 71.24 \\
\midrule

& ReSearch & 69.97 & 69.61 & 69.58 & 84.39 & 95.57 & 87.42 & 76.31 & 82.09 & 51.75 & 76.30 \\
\rowcolor{rowgray}& Search-R1 & 64.97 & 61.11 & 57.96 & 81.99 & 89.92 & 80.92 & 70.41 & 72.53 & 44.09 & 69.32 \\
& AEPO & 80.12 & 86.15 & 86.76 & 92.76 & 97.35 & 94.15 & 90.07 & 90.05 & 82.47 & 88.88 \\
\rowcolor{rowgray}& ARPO & 69.29 & 74.80 & 67.06 & 87.82 & 91.44 & 87.29 & 82.20 & 86.38 & 59.84 & 78.46 \\
& Mem1 & 74.71 & 70.02 & 63.50 & 85.85 & 93.50 & 85.87 & 73.62 & 74.20 & 48.96 & 74.47 \\
\rowcolor{rowgray}& KbPO & 65.20 & 51.80 & 61.50 & 56.30 & 71.90 & 54.70 & 62.30 & 67.80 & 32.60 & 57.99 \\
\rowcolor{avgblue}\multirow{-7}{*}{\makecell{RL-based\\Agentic RAG}}& \textbf{KbSD (Ours)} & \textbf{44.30} & \textbf{38.80} & \textbf{33.80} & \textbf{41.10} & \textbf{41.70} & \textbf{38.80} & \textbf{59.10} & \textbf{44.10} & \textbf{28.90} & \textbf{41.18} \\

\bottomrule
\end{tabular}
}
\caption{Comparison of \textbf{Unreliability Rate (\%)} (Lower is Better) across multi-hop and open-domain benchmarks. Score $= \text{F1} + \text{Refusal Rate}$; Unreliability Rate $= 1 - \text{Score}$.}
\label{tab:comparison}
\end{table*}

Table~\ref{tab:comparison} presents the Unreliability Rate across all benchmarks and model scales, where lower values indicate better reliability. We highlight several key findings.

\paragraph{Overall Performance.}
KbSD consistently achieves the lowest unreliability rate across all model scales, outperforming both knowledge-boundary-agnostic RL methods and the KbPO baseline that relies solely on sparse rewards. The improvement margin increases with model scale, suggesting that larger models better leverage the dense process-level supervision provided by self-distillation.

\paragraph{Knowledge Boundary Calibration.}
The largest gains appear on benchmarks requiring precise calibration. On TriviaQA and NQ, KbSD substantially outperforms Search-R1 and AEPO, with improvements stemming from KbSD's ability to refuse unknown queries rather than hallucinate. Compared to KbPO, which shares the same quadrant taxonomy but uses only sparse rewards, KbSD's dense distillation supervision provides clearer guidance on the reasoning process leading to calibrated decisions.

\paragraph{Out-of-Domain Generalization.}
Despite training only on HotpotQA and 2WikiMQA, KbSD generalizes well to unseen datasets including FRAMES and GAIA. This transferability suggests that knowledge boundary awareness, reinforced by process-level distillation, is a domain-agnostic skill rather than dataset-specific pattern fitting.

\paragraph{Comparison with RL-based Methods.}
KbSD shows consistent advantages over AEPO, ARPO, and MEM1. Critically, AEPO and ARPO exhibit instability at larger scales, likely resulting from reward hacking in outcome-only training. KbSD's quadrant-adaptive distillation loss prevents such exploitation by providing complementary dense supervision.

\subsection{Ablation Study}

We conduct ablation studies on Qwen2.5-3B to analyze the contribution of each design component.

\paragraph{Effect of Distillation Components.}
Table~\ref{tab:ablation} presents ablation results. Removing the distillation loss entirely ($-\mathcal{L}_{\text{distill}}$) degrades performance across all benchmarks, confirming that dense process-level supervision is essential. Replacing the quadrant-adaptive KL with a fixed forward KL (Fixed-FKL) or fixed reverse KL (Fixed-RKL) also hurts performance, particularly on benchmarks requiring diverse refusal behavior (TriviaQA) or concentrated integration reasoning (2WikiMQA), validating the necessity of adapting the divergence objective per knowledge state.

\begin{table}[!h]
\centering
\small
\setlength{\tabcolsep}{4pt}
\begin{tabular}{l|c|cc|c}
\toprule
\rowcolor{gray!20}
\textbf{Dataset} & \textbf{KbSD} & \textbf{$-\mathcal{L}_{\text{distill}}$} & \textbf{Fixed-FKL} & \textbf{Fixed-RKL} \\
\midrule
\rowcolor{rowgray}
TriviaQA & \textbf{34.80} & 39.90 & 39.70 & 50.50 \\
2WikiMQA & 45.90 & 70.50 & \textbf{44.00} & 68.90 \\
\rowcolor{rowgray}
HotpotQA & \textbf{41.60} & 57.60 & 45.50 & 59.90 \\
NQ       & \textbf{64.40} & 68.50 & 66.47 & 72.10 \\
\rowcolor{rowgray}
PopQA    & \textbf{45.60} & 73.10 & 50.00 & 70.20 \\
\midrule
Average  & \textbf{46.46} & 61.92 & 49.13 & 64.32 \\
\bottomrule
\end{tabular}
\caption{Ablation on distillation components (Unreliability Rate, \%, lower is better).}
\label{tab:ablation}
\end{table}

\paragraph{Effect of Quadrant Rebalancing.}
Proper calibration between Q1--Q4 reduces unreliability rates substantially, particularly on TriviaQA and NQ. Over-rewarding internal answering risks hallucination, while over-rewarding refusal sacrifices helpfulness. KbSD's quadrant-adaptive optimization finds the right balance across all knowledge states.

\subsection{Analysis}

To further validate KbSD's reliability, we analyze how its calibration behavior scales with model capacity and varies with task difficulty.

\paragraph{Effect of Model Scale.}
KbSD's reliability gains hold consistently across model scales. On Qwen2.5-3B, KbSD lowers the average unreliability rate from $64.32$ (KbPO) to $47.60$, and on Qwen2.5-7B from $57.99$ to $41.18$---an absolute reduction of roughly $17$ points in both cases. Notably, the larger model reaches a substantially lower absolute floor ($41.18$ vs.\ $47.60$), indicating that models with stronger parametric priors and reasoning capacity more effectively internalize the dense token-level demonstrations produced by the hinted teacher. This trend supports our central claim: the bottleneck addressed by KbSD is not model capacity but the \textit{sparsity} of process-level supervision, and providing dense, boundary-aware guidance pays off increasingly as the underlying policy becomes more capable.

\paragraph{Adaptive Refusal Behavior.}
KbSD exhibits difficulty-adaptive refusal rates that correlate with task complexity. On benchmarks where the model has sufficient knowledge, refusal rates remain low (2WikiMQA, PopQA, Bamboogle). On challenging benchmarks requiring complex reasoning or rare knowledge, refusal rates increase substantially (MuSiQue, FRAMES, GAIA). This adaptive pattern demonstrates that quadrant-based alignment, reinforced by self-distillation, successfully teaches the model to recognize its knowledge boundaries.

\paragraph{Correlation Between Difficulty and Calibration.}
We observe a clear negative correlation between F1 scores and refusal rates across benchmarks. This pattern confirms that KbSD has learned calibrated uncertainty: when both parametric knowledge and retrieval are insufficient, the model appropriately abstains rather than hallucinating. Compared to KbPO, KbSD's calibration gains are most pronounced in the challenging quadrants (Q3 and Q4), where sparse rewards are least informative---precisely the regime where dense distillation supervision provides the largest benefit.

\paragraph{Balancing Helpfulness and Reliability.}
Importantly, high refusal rates on difficult benchmarks do not compromise helpfulness on tractable tasks. On in-domain benchmarks (2WikiMQA, HotpotQA), KbSD maintains competitive F1 scores while keeping refusal rates low. This balance validates that quadrant-adaptive distillation successfully differentiates between scenarios requiring confident answers versus cautious abstention.

\subsection{Case Study}

We present two representative cases qualitatively illustrating KbSD's calibration behavior under contrasting knowledge states.

\begin{tcolorbox}[
  colback=avgblue!60!white,
  colframe=blue!40!black,
  fonttitle=\small\bfseries,
  title={Case 1 \quad Q4: Internal (Low Retrieval $+$ High Prior) \quad Bidirectional KL},
  left=5pt, right=5pt, top=4pt, bottom=4pt,
  boxrule=0.6pt
]
\small
\textbf{Question:} \textit{HMAS ``Platypus'' was a submarine depot ship operated by the Royal Australian Navy (RAN). HMAS ``AE1'' was an ``E''-class submarine of which organization?}\quad \textbf{GT:} Royal Australian Navy (RAN)

\smallskip
\textbf{Rollout Summary} (10 searches, Outcome 2.00, Retrieval 0.496):
The answer is implicitly stated in the question itself, yet the model performed 10 search iterations before converging. Early retrievals returned tangential information (torpedo specifications of AE1); the model eventually inferred through chained reasoning that HMAS AE1 belongs to the Royal Australian Navy.

\smallskip
\textbf{Analysis:} This Q4 case (high parametric prior, low retrieval quality) reveals a residual challenge: KbSD still occasionally over-relies on retrieval even when the parametric answer is accessible. The bidirectional KL objective guides the model toward the correct final answer, yet the excess search iterations indicate scope for further calibration of retrieval-efficiency under the Internal quadrant.
\end{tcolorbox}

\vspace{4pt}

\begin{tcolorbox}[
  colback=orange!8!white,
  colframe=myorange!80!black,
  fonttitle=\small\bfseries,
  title={Case 2 \quad Q3: Refusal (Low Retrieval $+$ Low Prior, Invalid Question) \quad Forward KL},
  left=5pt, right=5pt, top=4pt, bottom=4pt,
  boxrule=0.6pt
]
\small
\textbf{Question:} \textit{Which conductor, famous for leading orchestras using historical instruments, led a Verdi opera with a Moorish protagonist, collaborating with the Mahler Chamber Orchestra in the capital of the country where Frederick, heir to the Danish throne, was born and later married, though not in the same year?}\quad \textbf{GT:} Invalid question

\smallskip
\textbf{Rollout Summary} (4 searches, Outcome 0.50 IDK reward, Retrieval 0.363):
The model issued four targeted searches addressing each constraint in turn (conductor + historical instruments; Moorish protagonist; Mahler Chamber Orchestra; Danish royal lineage). Searches returned contradictory and incomplete evidence. The model initially proposed a candidate but immediately reconsidered, concluding: \textit{``Given the incomplete evidence, I do not have enough information to conclusively answer the question.''} Final answer: \textit{``I don't know.''}

\smallskip
\textbf{Analysis:} This Q3 case demonstrates the critical refusal capability reinforced by KbSD's forward KL objective. The question's multiple crossing constraints make it unanswerable; KbSD correctly identifies the evidence gap and abstains, earning the IDK reward (0.50) rather than hallucinating. This behavior---principled abstention in the face of insufficient evidence from both parametric memory and retrieval---is precisely the calibration that sparse-reward baselines such as KbPO struggle to achieve.
\end{tcolorbox}

\section{Conclusion}

We identify \textit{process-level reward sparsity} as a fundamental bottleneck in knowledge boundary calibration for agentic search: terminal rewards indicate \textit{which} behavior is correct but leave the intermediate reasoning trajectory unsupervised, which is especially limiting across the four heterogeneous knowledge-boundary quadrants. To address this, we propose \textbf{KbSD}, a knowledge-boundary-aware self-distillation framework. KbSD constructs a hint-augmented \textit{teacher} that is architecturally identical to the \textit{student} but additionally conditioned on privileged boundary signals---parametric certainty, semantic stability, retrieval quality, and the target quadrant---and distills its calibrated reasoning into the hint-free student. This information-asymmetric design converts sparse behavioral labels into dense, token-level process supervision without requiring a larger external teacher. To respect the distinct optimization geometry of each quadrant, KbSD further employs a quadrant-adaptive objective---reverse KL for concentrated integration, forward KL for diverse refusal, and Pareto-optimal bidirectional KL for the precision--coverage trade-off in the asymmetric quadrants---jointly optimized with GRPO.

Experiments on nine benchmarks and two model scales show that KbSD consistently reduces unreliability over strong baselines, cutting the average unreliability rate from $64.32$ to $47.60$ on Qwen2.5-3B and from $57.99$ to $41.18$ on Qwen2.5-7B, with the largest gains concentrated in the challenging quadrants where sparse rewards are least informative. Ablations confirm that both the distillation loss and its quadrant-adaptive form are essential, and our analysis shows that the gains grow with model scale. As the case studies illustrate, a residual challenge remains in calibrating retrieval efficiency under the Internal quadrant, where the agent can still over-search despite an accessible parametric answer. We leave a tighter coupling between knowledge-boundary awareness and retrieval budgeting, as well as extensions to larger models and open-ended agentic settings, to future work.

\bibliography{aaai2026}

\end{document}